\documentclass[sigconf]{acmart}
\AtBeginDocument{%
  }

\copyrightyear{2026}
\acmYear{2026}
\setcopyright{cc}
\setcctype{by}
\acmConference[KDD '26]{Proceedings of the 32nd ACM SIGKDD Conference on Knowledge Discovery and Data Mining V.1}{August 09--13, 2026}{Jeju Island, Korea}
\acmBooktitle{Proceedings of the 32nd ACM SIGKDD Conference on Knowledge Discovery and Data Mining V.1 (KDD '26), August 09--13, 2026, Jeju Island, Republic of Korea}
\acmPrice{}
\acmDOI{10.1145/3770854.3783953}
\acmISBN{979-8-4007-2258-5/2026/08}

\usepackage{booktabs}   
\usepackage{tabularx}   
\usepackage{multirow}   
\usepackage{graphicx}
\usepackage{subcaption}
\usepackage{enumitem}

\usepackage{tcolorbox}
\tcbset{
  promptstyle/.style={
    colback=gray!15,   
    colframe=gray!60,  
    boxrule=0.4pt,     
    arc=2pt,           
    left=4pt,right=4pt,top=4pt,bottom=4pt,  
    fonttitle=\bfseries,
    width=\linewidth,  
  }
}
\newtcolorbox{promptblock}[1][]{promptstyle,#1}

\begin{document}

\title{From Clicks to Preference: A Multi-stage Alignment Framework for Generative Query Suggestion in Conversational System}



\author{Junhao Yin}
\affiliation{%
  \institution{Bytedance}
  \city{Shanghai}
  \country{China}}
\email{yinjunhao@bytedance.com}

\author{Haolin Wang}
\affiliation{%
  \institution{Bytedance}
  \city{Beijing}
  \country{China}}
\email{wanghaolin.11@bytedance.com}

\author{Peng Bao}
\affiliation{%
  \institution{Bytedance}
  \city{Beijing}
  \country{China}}
\email{pengbao7598@gmail.com}

\author{Ju Xu}
\affiliation{%
  \institution{Bytedance}
  \city{Beijing}
  \country{China}}
\email{yufeng.1016@bytedance.com}

\author{Yongliang Wang}
\affiliation{%
  \institution{Bytedance}
  \city{Beijing}
  \country{China}}
\email{yongliang.wyl@bytedance.com}


\begin{abstract}
Generative query suggestion using large language models offers a powerful way to enhance conversational systems, but aligning outputs with nuanced user preferences remains a critical challenge. To address this, we introduce a multi-stage framework designed for progressive alignment between the generation policy and user intent. Our pipeline begins with prompt engineering as a cold-start strategy, followed by the Supervised Fine-Tuning stage, in which we introduce a distillation method on click logs to create a robust foundational model. To better model user preferences while capturing their inherent uncertainty, we develop a Gaussian Reward Model (GaRM) that represents user preferences as probability distributions rather than point estimates. Finally, we employ reinforcement learning to align the generation policy with these preferences, guided by a composite reward function that integrates GaRM with auxiliary heuristics to mitigate reward hacking. To maintain training stability, this process is enhanced by a novel out-of-distribution regularization method and a two-stage reward fusion technique. Extensive experiments demonstrate that our framework significantly outperforms baselines on both automatic and human evaluations and yields a 34\% relative increase in user engagement as measured by click-through rate in live A/B tests.

\end{abstract}




\maketitle
\begin{figure}[t]
    \centering
    \includegraphics[width=1.0\linewidth]{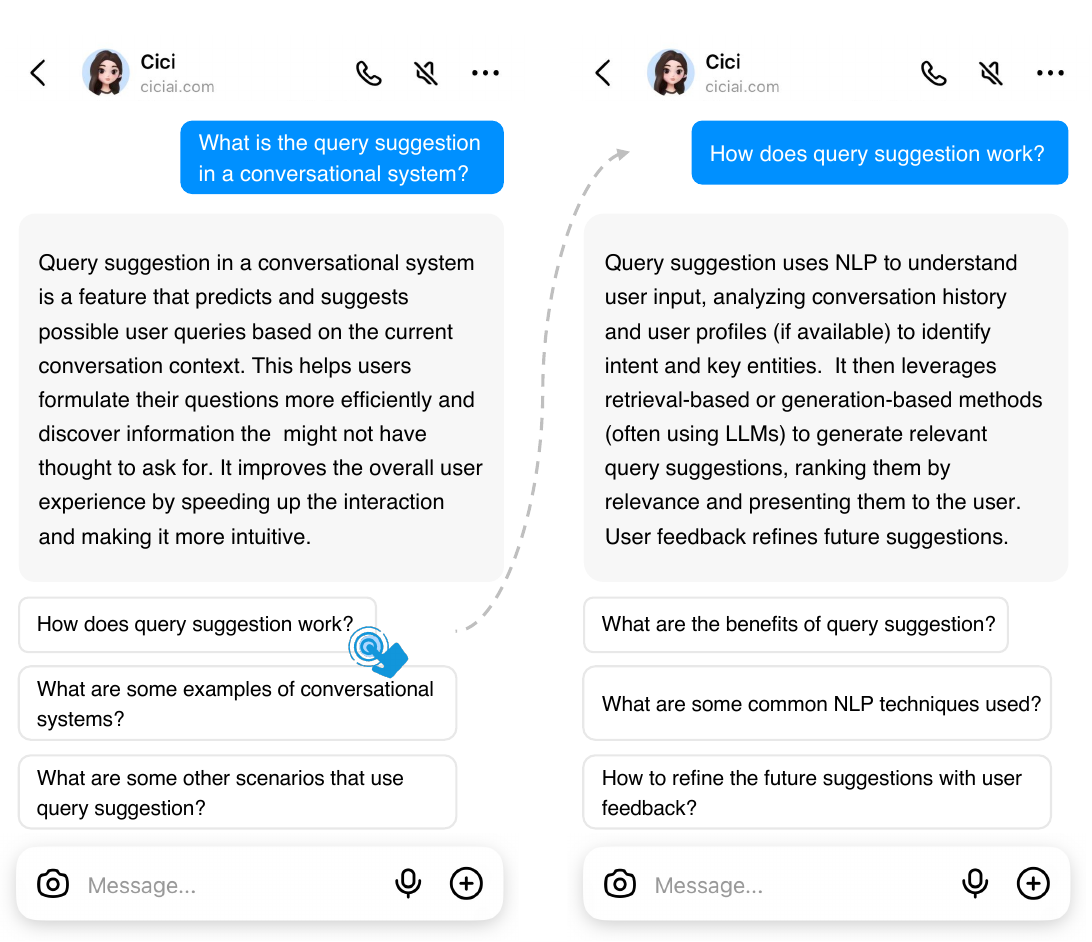}
    \caption{An example of QS in a conversational system.}
    \label{fig:cici_example}
\end{figure}

\section{Introduction}

The proliferation of conversational systems, including chatbots and virtual assistants, has significantly reshaped the landscape of human-computer interaction~\citep{gao2018neural}. These systems are rapidly moving beyond niche applications to become ubiquitous platforms for information access, task completion, and customer support~\citep{shum2018eliza}. 
With this evolving landscape, query suggestion (QS) has emerged as a cornerstone of effective conversational user experience. As illustrated in Figure~\ref{fig:cici_example}, its primary function is to proactively recommend relevant queries based on the conversational context, making the interaction faster and more satisfying. 
Ideally, an intelligent QS module should not merely autocomplete user queries, but actively lead the conversation toward a success outcome.

\begin{figure*}[!htbp]
\centerline{\includegraphics[width=1.04\linewidth]{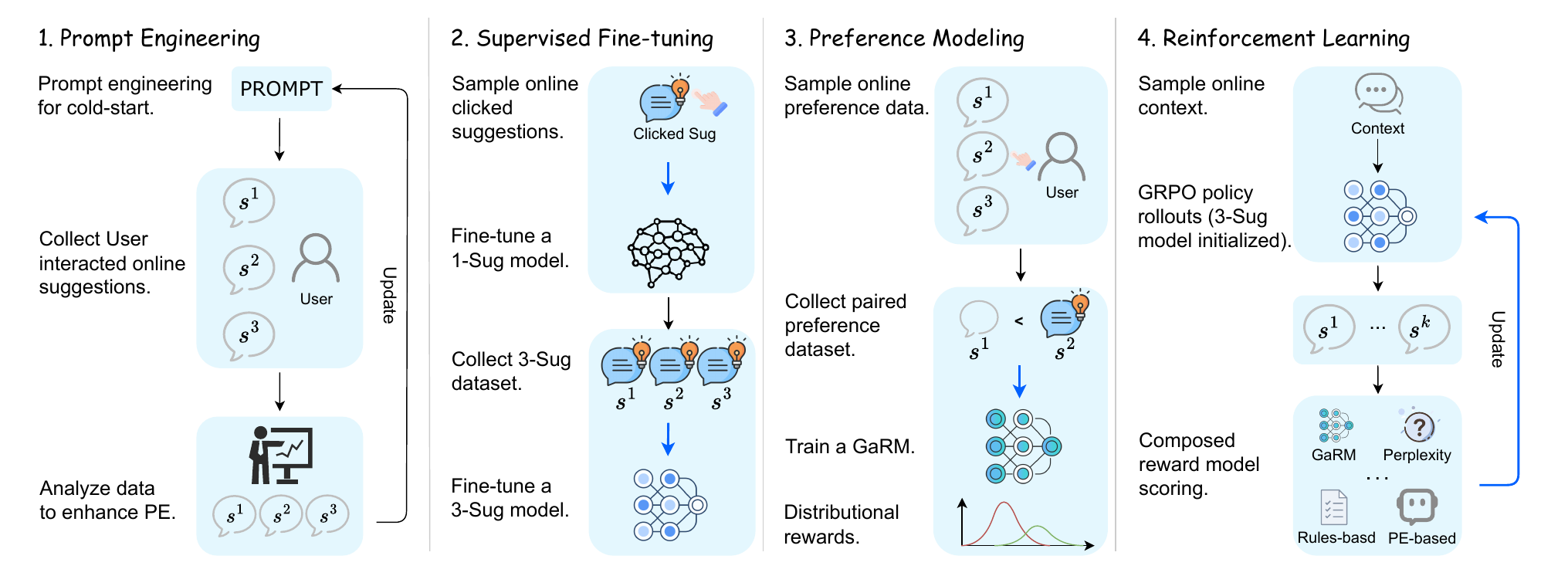}}
    \caption{An overview of proposed four-stage framework for suggestion generation.} 
    \label{fig:framework}
\end{figure*}

Traditional methods in QS, including collaborative filtering~\cite{schafer2007collaborative} and session-based recommendations~\cite{chen2021collaborative, ludewig2021empirical, ludewig2018evaluation}, operate on a retrieval-based paradigm, selecting suggestions from a finite candidate pool of previously observed queries. This design makes them inherently unable to generate novel, contextually specific suggestions and leaves them vulnerable to chronic issues like data sparsity and the cold-start problem~\cite{fayyaz2020recommendation}.
The advent of Large Language Models (LLMs)~\cite{achiam2023gpt, gemma_2025, guo2025deepseek} has catalyzed a paradigm shift from retrieval-based matching to semantic generation, directly addressing these foundational constraints with their superior contextual understanding and world knowledge~\cite{wang2023zero, he2023large, wang2024multimodal}.

While LLMs solve many of the long-standing problems in traditional methods, their generative power introduces a formidable new challenge: aligning their output with true, often latent, user preferences. This raises two critical research questions: 1) How can user preferences be accurately modeled from noisy, implicit signals such as click logs? and 2) How can this preference model be used to steer a generative model towards robust alignment with user intent?
To address this alignment gap, we propose a holistic, multi-stage framework, as illustrated in Figure~\ref{fig:framework}, which systematically progresses from coarse behavioral signals toward a nuanced and robust model of true user preference.

The initial stage of our framework addresses the cold-start problem via an elaborate \textbf{Prompt Engineering} (PE). Here, prompts are meticulously engineered to elicit suggestions exhibiting both high relevance and diversity, a strategy informed by empirical findings from our preliminary studies. 
Following this, we undertake a \textbf{Supervised Fine-Tuning} (SFT) phase to distill observed user behavior into the model, primarily by fine-tuning on real-world click logs. Specifically, we introduce a progressive data construction strategy that uses a powerful offline model to synthesize high-quality suggestion sets, ensuring both the quality of individual suggestions and the diversity within the group. Nevertheless, a sole reliance on click data for SFT presents inherent constraints. It captures only positive feedback, neglecting the information from unclicked negative samples. Furthermore, a click often signifies a sufficient but potentially suboptimal choice. An SFT model is therefore confined to mimicking this imperfect behavior, limiting its capacity to improve upon it. This motivates the transition to a reward model, which is designed to learn the latent preference from user behavior instead of merely replicating observed actions.


In the QS context, an additional challenge is that user choice is non-deterministic and noisy, making a reward model based on a deterministic score imprecise.
To this end, we propose the \textbf{Gaussian Reward Model} (GaRM), which conceptualizes preference scores not as deterministic scalars but as probabilistic distributions. Critically, we derive an analytically tractable loss function with variance regularization for GaRM, overcoming the inefficiency and instability of prior probabilistic approaches~\citep{sun2025probabilistic} and making it robust to the inherent noise in preference data.


The final stage of our framework employs \textbf{Reinforcement Learning} (RL) to directly optimize the generative policy against our refined preference understanding. To ensure stable policy updates and mitigate the significant risk of reward hacking, wherein the model learns to exploit artifacts in the reward signal, we design a composite reward function. This function avoids reliance on any single, brittle signal and integrates the probabilistic scores from GaRM with auxiliary signals, including rule-based heuristics and prompt-based quality metrics. Furthermore, we introduce a regularization term based on logged perplexity from the reference model. We theoretically prove that this method is equivalent to constraining the out-of-distribution (OOD) extent of the reward model, which effectively prevents the RL policy from exploring regions where the reward model cannot provide accurate scores. Finally, to blend these diverse signals, we introduce a two-stage fusion process. This process first determines initial weights using logistic regression and subsequently refines this balance through a Pareto-guided search to achieve stable, multi-objective gains.


In summary, this paper proposes a novel, multi-stage framework that systematically progresses from coarse behavioral signals toward a nuanced and robust model of true user preference. Within this framework, our key innovations are:
\vspace{-2pt}
\begin{itemize}[leftmargin=*]
\item We introduce a strategic data curation method for SFT that leverages click logs while employing a novel sampling scheme to explicitly enhance suggestion diversity.
\item We propose the GaRM as a form of probabilistic preference learning to capture the inherent uncertainty in user intent.
\item We develop a robust reinforcement learning methodology, featuring a composite reward function, a theoretically-grounded regularization to constrain OOD exploration, and a two-stage fusion process to balance reward signals.
\item We demonstrate the efficacy of our framework through extensive offline experiments and online A/B testing, showing significant improvements over baselines in both automatic and human evaluation metrics.
\end{itemize}

\section{Related Work}
\subsection{Query Suggestion in Search Engine}
Query Suggestion (QS) has been extensively studied within the context of traditional search engine, where its primary goal is to assist users by recommending queries based on historical usage patterns. Early QS research relied on mining query co-occurrence in logs, click-through statistics, and simple probabilistic models such as n-grams or Markov chains~\cite{baeza2004query, cao2008context}. While effective for popular or repetitive queries, such approaches fall short in open-domain conversations due to limited semantic understanding and inability to handle unseen inputs. To overcome these limitations, researches introduced neural network-based methods, particularly recurrent neural networks and attention-based models~\cite{sordoni2015hierarchical, mustar2020using, mustar2021study, dehghani2017learning}, to better capture sequential dependencies in session data and generalize across diverse queries. However, these systems are still constrained by a fixed set of candidate queries and require extensive offline training data.

\subsection{Query Suggestion in Conversational Systems}
In contrast to traditional search QS, conversational QS aims to proactively guide the conversation toward more fruitful, engaging, or task-completing directions, a concept termed "conversation-leading" suggestions~\cite{rosset2020leading}. This capability has become increasingly central in large-scale conversational AI systems like ChatGPT~\cite{ouyang2022training}, Claude~\cite{bai2022constitutional}, and Doubao~\cite{seed2025th}, where users often rely on system-initiated suggestions to navigate complex, multi-turn interactions. Rosset \textit{et al.}~\cite{rosset2020leading} pioneered the notion of conversation-leading query suggestions—informative prompts that anticipate user needs rather than merely continuing the current thought. Wang \textit{et al.}~\cite{wang2023zero} further demonstrated the efficacy of zero-shot LLMs in generating high-quality query continuations and Li \textit{et al.}~\cite{Li2025ProactiveGO} train a language model to enhance search chat-bot. Such generative approaches are particularly valuable in cold-start or exploratory settings, but they also introduce challenges. Without proper alignment, LLMs may hallucinate content, repeat generic suggestions, or deviate from useful dialogue trajectories.

\subsection{Human Preference Alignment}
Aligning generative models with nuanced human preferences is a central challenge in modern AI. The dominant paradigm for this is Reinforcement Learning from Human Feedback (RLHF), which consists of three main stages: SFT, reward model (RM) training, and RL-based policy optimization. The core idea of learning from human preferences was pioneered by Christiano et al.~\cite{christiano2017deep}, and the full RLHF pipeline was popularized by its successful application to large-scale language models like InstructGPT~\cite{ouyang2022training}. In the context of QS, the primary challenge is defining a suitable preference signal. Consequently, research has focused on using implicit user feedback, such as clicks, as a proxy for preference~\cite{min2025prompting}. For instance, Min \textit{et al.} designed a system that first train a click-through rate (CTR) predictor to act as a reward model and then use it to guide the generative model toward suggestions that are more likely to be clicked~\cite{min2025prompting}. Furthermore, the authors used the predicted CTR to weight the importance of preference pairs in the Direct Preference Optimization (DPO) loss function and combined it with a diversity-aware regularization term to ensure that the suggestions remain varied~\cite{min2025ctr}. However, relying on CTR models as preference proxies presents significant challenges. Primarily, such models tend to associate clicks with the input context, rather than modeling the relative quality differences between candidate suggestions. Furthermore, user clicks are stochastic by nature, but a simple CTR model cannot represent this probabilistic uncertainty, treating each signal as a deterministic event. In response to these issues, we propose a probabilistic, pairwise GaRM to explicitly model the uncertainty in user preference. We then employ a full RL stage, complete with a composite reward and novel regularization, to achieve a more robust and holistic alignment with true user satisfaction.

\section{Method}
\subsection{Problem Formulation}
Let \( h \) denote the historical interaction between the user and the AI system. The generation model \( \pi_\theta \) takes \( h \) as input and outputs three different query suggestions in the form of an ordered list, denoted as \( \langle s^1, s^2, s^3 \rangle \). Under this setting, our objective is to maximize the expected CTR. Formally, the optimization goal can be defined as:
\begin{equation}
    \underset{\theta}{\max}\ \mathbb{E}_{h\sim\mathcal{H}}\mathbb{E}_{\langle s^1, s^2,s^3\rangle\sim\pi_\theta(\cdot|h)}\text{CTR}(\langle s^1, s^2, s^3\rangle),
\end{equation}
where $\text{CTR}(\langle s^1, s^2, s^3\rangle)$ denotes the probability that the user clicks on any of the suggestions in the list.

\subsection{System Architecture}
\begin{figure*}[t]
\centerline{\includegraphics[width=0.82\linewidth]{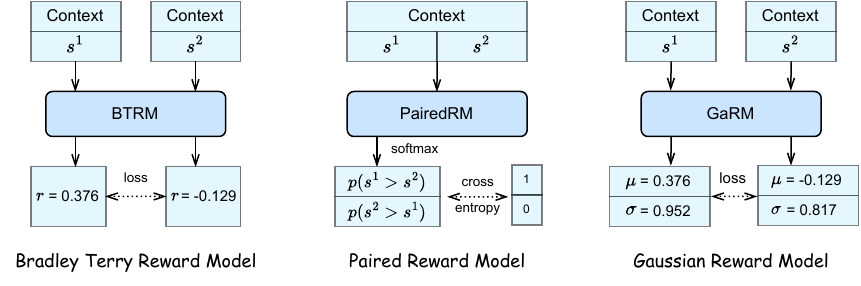}}
    \caption{Comparison of different reward models.}
    \label{fig:reward_models}
\end{figure*}

The architecture of our framework, illustrated in~\autoref{fig:framework}, details the comprehensive pipeline for bootstrapping and iteratively improving our query suggestion model. The pipeline consists of the following key phases:
\begin{enumerate}[leftmargin=*]
    \item \textbf{Prompt Engineering for Cold-start}: We leverage prompt engineering (PE) to power the initial launch of the system. The sole purpose of this phase is to generate initial suggestions and collect the first batch of real-world user click data.
    \item \textbf{Supervised fine-tuning}: With the collected click data, we perform a two-stage SFT process to adapt the base model to our task. First, we fine-tune a "single-suggestion model" whose primary goal is to generate one high-quality query. Next, we use this model to curate a more complex dataset, which is then used to further fine-tune the final SFT model to generate a list of three suggestions.
    \item \textbf{Preference modeling}: In this phase, the collected click data is used to train a RM that learns to distinguish between clicked and unclicked suggestions, thereby accurately modeling user preferences. 
    \item \textbf{Reinforcement learning}: Finally, RL is applied to optimize SFT model, utilizing the RM in combination with other heuristic-based rewards to further enhance suggestion quality.
\end{enumerate}

\subsection{Prompt Engineering}
The initial stage of our framework addresses cold-start generation through principled prompt engineering. In a conversational system, a suggested query is deemed successful if its execution leads to a response that either (1) directly resolves the user's underlying intent or (2) provides a natural and valuable extension to the current conversational path. Complementing these positive objectives, we define a comprehensive set of negative constraints spanning issues from factual inaccuracies to conversational missteps (see Appendix \ref{sec:appendix_evaluation_standard} for a full specification) to ensure quality and avoid common interactional failures. These criteria are then systematically translated into a structured prompt, which is iteratively refined based on live online data. Our prompt template includes history contexts and evaluation criteria, detailed in Appendix \ref{sec:prompt_template}. The deployment of this prompt-driven system is thus foundational, serving not only as an effective cold-start solution but also as the data collection engine for the subsequent SFT and preference modeling stages.



\subsection{Supervised Fine-tuning}
A primary challenge in SFT for query suggestion is the noisy nature of click data: a click on one suggestion provides no quality signal for other co-displayed suggestions. To address this, we propose a progressive data construction methodology to create a high-quality, multi-suggestion training dataset from this noisy foundation, focusing on both the quality of individual suggestions and the diversity within each group. Specifically, our methodology consists of two primary stages.

\textbf{High-Quality Candidate Synthesis}. We begin by training a powerful, large-scale "teacher" model (e.g., Gemma3-27B~\citep{gemma_2025}, Qwen3-32B~\citep{qwen3technicalreport}) on a cleaned, single-suggestion dataset. Freed from online inference constraints, this model is optimized solely for generation quality. We then use this teacher model to perform parallel sampling for each source dialogue context, generating a large candidate pool of high-quality individual suggestions.

\textbf{Diversity-Aware Assembly}. Next, we assemble the final three-suggestion training instances through a rigorous filtering process designed to maximize intra-group diversity. We employ a multi-faceted approach to remove redundant candidates, using signals such as embedding similarity and n-gram overlap. The filtered, high-quality suggestions are then grouped to form the final, diverse training dataset for our main SFT model.

\subsection{Preference Modeling}
This section details the models developed to modeling user preferences, which serve as the RM for the subsequent RL phase. We first describe our dataset curation process, followed by different modeling approaches.

\textbf{Dataset Curation.} We adopt a standard triplet format to construct the training data for the reward models. Each data point is represented as \( \langle h, s_w, s_l \rangle \), corresponding to the historical dialogue context, a user-preferred suggestion, and a user-disfavored suggestion, respectively. A straightforward approach to selecting positive and negative suggestions is to treat the clicked suggestion among the three as positive and the other two as negative. However, we observed that users tend to click on the first suggestion more often due to positional bias, which may distort the modeling of true user preferences. To address this, we filter the data by only retaining samples in which the clicked suggestion is not in the first position and treat the earlier-positioned suggestion as the negative sample. More details on the data construction process can be found in Appendix~\ref{sec:appendix_datasets}.

We now describe three different reward modeling approaches: the Bradley-Terry Reward Model (BTRM), the Paired Reward Model (PairedRM), and the Gaussian Reward Model (GaRM). Their conceptual workflows are illustrated in Figure~\ref{fig:reward_models}.

\textbf{Bradley-Terry Reward Model.} This is a classical modeling approach that maps each suggestion and its context into a scalar score \( r_\phi(s; h) \), which we use as a baseline in this work \citep{ouyang2022training, wang2024secrets}. The loss function is defined as:
\begin{equation}
\mathcal{L}_{\phi} = -\mathbb{E}_{\langle h, s_w, s_l \rangle} \left[ \log\ \text{sigmoid}\left(r_\phi(s_w; h) - r_\phi(s_l; h)\right) \right].
\end{equation}

\textbf{Paired Reward Model.} We also evaluate an improved approach by feeding both \( s_w \) and \( s_l \) along with context \( h \) into the reward model, which directly outputs the probability that \( s_w \) is preferred over \( s_l \):
\begin{equation}
\mathcal{L}_{\phi} = -\mathbb{E}_{\langle h, s_w, s_l \rangle} \left[ \log r_{\phi}(s_w, s_l; h) \right].
\end{equation}
This approach is known as PairedRM in previous literature~\citep{pairedrm1, pairedrm2, pairdrm3}. Further implementation details are provided in Appendix~\ref{sec:appendix_pairedrm}.

\textbf{Gaussian Reward Model.} Due to the inherent variability in user preferences, the dataset of suggestion preferences contains significant noise that is difficult to eliminate through filtering alone. Therefore, it is crucial to design a more robust reward model at the algorithmic level. To this end, we propose modeling user preference for a suggestion as a probability distribution rather than a scalar value, as in BTRM. Inspired by and improving upon PURM~\citep{sun2025probabilistic}, we model preferences as a parameterized distribution \( \mathcal{N}(\mu, \sigma^2) \). In the original PURM formulation, the optimization objective is to maximize:
\begin{equation}
p(s_w > s_l; h) 
=\int\text{sigmoid}(z) \, \mathcal{N}\left(z \mid \mu_w - \mu_l, 
{\sigma_w^2 + \sigma_l^2} \right) \, dz,
\label{eq:purm}
\end{equation}
where \((\mu_w, \sigma_w) = r_{\phi}(s_w; h),  (\mu_l, \sigma_l) = r_{\phi}(s_l; h)\). To compute the integral in Equation~\ref{eq:purm}, the original paper employs a Monte Carlo sampling approach. However, we identify two key limitations with this method. First, Monte Carlo estimation is computationally inefficient and may lead to unstable or inaccurate approximations, which we prove in Appendix~\ref{sec:appendix_garm_loss}. Second, we observe that this loss function tends to encourage the model to output larger values of \( \sigma \), which results in unstable training dynamics. To address these issues, we propose an analytically tractable loss function:
\begin{equation}
\begin{aligned}
\mathcal{L}_{\phi} = &- \mathbb{E}_{\langle h, s_w, s_l \rangle} [\log \text{sigmoid} \left( \frac{\mu_w - \mu_l}{\sqrt{1 + \frac{\pi}{8} \left(\sigma_w^2 + \sigma_l^2\right)}} \right)
\\ &+ \lambda \left( \sigma_w^2 - 2\ln \sigma_w + \sigma_l^2 - 2\ln \sigma_l \right)].
\label{eq:garm}
\end{aligned}
\end{equation}
The first term in the loss function provides a closed-form approximation to Equation~\ref{eq:purm}, while the second term regularizes the variance outputs, encouraging \( \sigma \) values to remain close to 1. This regularization helps stabilize training. A detailed derivation of the proposed loss and its approximation can be found in Appendix~\ref{sec:appendix_garm_loss}.

The application of our proposed GaRM and the original PURM also differs significantly during the RL phase. In the original PURM, the authors utilize the Bhattacharyya Coefficient (BC) to measure the confidence in distinguishing between positive ($s_w$) and negative ($s_l$) samples, defined as:
\begin{equation}
    \text{BC} = \sqrt{\frac{2\sigma_w\sigma_l}{\sigma_w^2 + \sigma_l^2}} \exp\left(-\frac{(\mu_w - \mu_l)^2}{4(\sigma_w^2 + \sigma_l^2)}\right).
\end{equation}
However, this metric is challenging to apply directly in the RL phase as it depends on two distinct suggestions, whereas the RL framework requires an independent reward for each individual suggestion. To address this, we introduce an enhancement by defining a lower bound for the Bhattacharyya distance, which we term the Uncertainty Lower Bound (ULB). By maximizing this ULB during the RL phase, we aim to increase the separability of the sample distributions, thereby improving the confidence of the distinction. The ULB is formulated via the Bhattacharyya distance ($D_B = -\log \text{BC}$) and its established lower bound:
\begin{equation}
    D_B \ge \frac{(\mu_w - \mu_l)^2}{4(\sigma_w + \sigma_l)^2}.
    \label{eq:ulb}
\end{equation}
The rationale for the ULB is discussed in our experiments, with its full derivation provided in Appendix \ref{sec:appendix_ulb}. Furthermore, given that the term $(\mu_w - \mu_l)^2$ is bounded in practice, we simplify the objective for the RL agent. Specifically, we incorporate the negative standard deviation ($-\sigma$) as a component of the reward function. Intuitively, this encourages the RL policy to generate samples that not only achieve high scores but also have low uncertainty, corresponding to predictions where the reward model is most confident.

\subsection{Reinforcement Learning}
During the RL stage, we find that relying solely on a trained reward model is far from sufficient. This is primarily because, in this domain, users exhibit distinct preferences for suggestions that convey similar meanings but differ slightly in phrasing. Consequently, a trained reward model faces two significant challenges. First, it frequently encounters unseen samples, leading to out-of-distribution (OOD) issues that significantly degrade its accuracy. Second, the reward model may overly focus on stylistic features, thereby neglecting critical aspects such as coherence with recommended terms and historical context. A detailed case study on this topic is provided in Appendix~\ref{sec:appendix_case_RM}.

To address these challenges, we propose leveraging multiple reward signals and jointly optimizing them during reinforcement learning. We categorize these reward signals into four types. \textbf{Rule-based reward} refers to programmatically defined scoring functions applied to generated texts, including metrics such as diversity checks and language consistency. \textbf{Prompt-based reward} uses API calls to LLMs, where we provide instructions defining the scoring criteria and let the LLM assign a score. \textbf{Regularization reward} is introduced to discourage the fine-tuned model from deviating excessively from the original model; in this work, we use the logged perplexity (PPL) as a regularization term. \textbf{Neural network-based reward} includes the three types of reward models trained in the previous section, along with the pretrained reward model provided by Skywork~\citep{liu2024skywork}. 
We briefly discuss several types below, and more information can be found in Appendix~\ref{sec:appendix_aux_rewards}.

\subsubsection{Rule-based rewards.} Aligning with recent findings, which demonstrate that rule-based 'checklists' are an effective strategy for reward engineering~\citep{viswanathan2025checklists}, we employ several verifiable rewards to ensure more controllable and desirable model behavior. These include:

\textbf{Format Reward.} This reward verifies that the generated output adheres to predefined formatting rules. Primarily, it checks that exactly three suggestions are generated as required.

\textbf{Length Reward.} To encourage conciseness, we penalize suggestions that are excessively long. We set a soft limit for each suggestion at a length of 12 words. If a suggestion surpasses this limit, a penalty is applied that scales linearly with the number of excess words~\citep{yu2025dapo}.

\textbf{Language Consistency Reward.} This reward ensures that the language of the generated suggestions is consistent with the language of the user's original query. A penalty is applied if an inconsistency is detected.

\textbf{Diversity Reward.} To promote a varied set of outputs, this reward measures the 1-gram similarity among the three generated suggestions. The reward is inversely proportional to the similarity score, meaning lower similarity yields a higher diversity reward.

\textbf{Safety Reward.} For inputs identified as having potential safety issues, the model is required to decline the request by outputting a specific ``Unsafe'' token. A significant penalty is applied if the model provides any other response to such inputs.

\subsubsection{Regularization with Logged Perplexity.} As discussed above, one of the critical challenges during the RL stage is that the generative model may explore OOD regions where reward models are unreliable—leading to reward hacking. Thus, identifying whether a generated sample lies in the OOD region of a reward model is a key technical problem. Formally, taking the BTRM as an example, for a given sample $\langle h, s \rangle$, determining whether it is OOD for the reward model reduces to modeling the joint probability $p(s; h) = p(s \mid h) \cdot p(h)$. Here, $p(h)$ represents the prior over the historical context, which is largely governed by the user's interest distribution and changes slowly over time. In most cases, this joint distribution is intractable. However, in our setup, it is tractable because the reward model is trained on data generated by a reference model $\pi_{\text{ref}}$. Therefore:
\begin{equation}
p(s \mid h) = \pi_{\text{ref}}(s \mid h) \propto \log \text{ppl}_{\pi_{\text{ref}}}(s \mid h).
\end{equation}
Thus, the logged perplexity under the reference model serves as an approximate indicator of whether the sample is in-distribution for the reward model. 


\subsubsection{Reward Fusion}

In the previous section, we introduce various reward signals to stabilize updates during RL. In practice, these diverse reward signals are combined into a single reward value through a weighted average for subsequent RL optimization. The key question is how to determine the weights for this weighted average. We propose a two-stage approach to address this.

First, we employ logistic regression to compute initial weight values. Specifically, let \( r_{w}^j \) denotes the reward value for the \( j \)-th reward signal corresponding to a high-quality suggestion, and \( r_{l}^j \) denote the reward value for a low-quality suggestion. The goal is to find a set of parameters \( \mathbf{w} = \{w_j\} \) such that, for as many samples as possible, the following condition holds:
\begin{equation}
\sum_{j} w_j (r_{w}^j - r_{l}^j) > 0.
\end{equation}
This is essentially a classification problem, where the objective is to separate all samples onto one side of a hyperplane. Thus, we solve this using standard logistic regression, defined as:
\begin{equation}
\mathcal{L} = -\mathbb{E}_{\langle h, s_w, s_l \rangle} \left[ \log\ \text{sigmoid}\left(\sum_{j} w_j (r_{w}^j - r_{l}^j)\right) \right] + \lambda \|\mathbf{w}\|_2^2,
\end{equation}
where an L2 regularization term is included to ensure the weights remain balanced, preventing the overall reward from overly depending on any single reward signal and avoiding degradation.

Second, we propose a heuristic parameter tuning strategy. Using only the weights obtained from the first stage, we observe that during RL training, while the weighted average reward may increase, certain reward components consistently decrease, while others rise rapidly. Moreover, the resulting model performance is often suboptimal. This occurs because the weights computed in the first stage are conditioned on the dataset and tied to the generative model. Once the model’s parameters change during training, these weights cease to be optimal. To address this, we introduce the concept of Pareto optimality, ensuring that the improvement of any reward component does not lead to the degradation of others. Specifically, starting from the initial weights, we manually fine-tune them as follows: if a reward component decreases during RL, we increase its weight; if a single reward component dominates the overall reward increase, we reduce its weight. This process is iterated until all reward components exhibit a stable upward trend during RL training.

\subsubsection{RL Optimization}

We adopt Group Relative Policy Optimization (GRPO)~\citep{shao2024deepseekmath} to maximize the reward signal. GRPO optimizes the policy \(\pi_\theta\) by maximizing a penalized objective that balances reward maximization against policy divergence. For each sample \(i\) with input \(h_i\), the policy generates an output \(s_i\). The update leverages a reward signal \(R(h_i, s_i)\) and constrains the policy shift using the Kullback-Leibler divergence. The objective is formulated as:  

\begin{equation}
\begin{aligned}
   \theta_{\text{new}} = \arg\max_{\theta} & \Bigg\{ \mathbb{E}_{(h, s) \sim \pi_{\theta_{\text{old}}}} \left[ \frac{\pi_{\theta}(s | h)}{\pi_{\theta_{\text{old}}}(s | h)} R(h, s) \right] \\
   & - \beta \cdot \mathbb{E}_{h} \left[ \text{KL}\Big( \pi_{\theta}(\cdot | h) \parallel \pi_{\theta_{\text{old}}}(\cdot | h) \Big) \right] \Bigg\} .
\end{aligned}
\end{equation}
Here, \(\beta > 0\) is a regularization coefficient that controls the penalty strength. The first term importance-samples rewards under the updated policy \(\pi_\theta\), while the second term penalizes deviations from the previous policy \(\pi_{\theta_{\text{old}}}\). Expectations are approximated via Monte Carlo sampling over a batch of data.

\section{Experiments}
\subsection{Setup}
Our framework is implemented and evaluated within a large-scale, production conversational AI system--Cici. To ensure the models are trained and tested on representative data, all datasets are sampled from anonymized, real-world user interactions with the live system. We adopt the Qwen3-30B-A3B foundation model~\citep{qwen3technicalreport} for its state-of-the-art performance and computational efficiency—two key requirements for deployment in a production environment. 

\textbf{Training Datasets.} Our training pipeline leverages three distinct datasets. First, a SFT dataset of 100,000 high-quality samples, balanced across various user intents, is used for initial model adaptation. Second, the reward model is trained on a large-scale dataset of 950,000 preference pairs, each consisting of a user-clicked ("chosen") and an ignored ("rejected") suggestion derived from real interactions. Third, the RL phase utilizes a separate, disjoint set of 100,000 prompts, which includes a subset of "unsafe" queries to teach the model refusal capabilities. A detailed description of each dataset's curation and cleaning process is provided in Appendix~\ref{sec:appendix_datasets}. For comparative analysis, we include Rejection Sampling Fine-Tuning (RFT), a method training samples selected by the reward model. Implementation details of RFT are listed in Appendix~\ref{sec:appendix_rft}.

\textbf{Test Datasets.} For comprehensive evaluation, we construct two distinct test sets. The primary \textit{Suggestion Generation Test Set}, containing 100 diverse contexts, is used to assess the generative quality of all models (PE, SFT, RFT, and RL). To evaluate safety alignment, we construct a dedicated \textit{Safety Test Set} comprising 200 labeled unsafe contexts, measuring the model's ability to adhere to safety protocols by correctly suppressing suggestions in response to inappropriate inputs.

\subsection{Metrics}

\subsubsection{Offline evaluation.} We assess the quality of each generated suggestion group (containing up to three items) by assigning an integer score from 0 to 3, corresponding to the number of useful suggestions. Scoring is conducted through a hybrid process: initial annotations are performed by gpt-o3-mini~\citep{openai_o3_mini_system_card}, followed by a manual review by six human experts. Our offline metric, Good Same Bad (GSB), is then calculated as the aggregate score difference between a candidate model and a baseline across the test set.

\subsubsection{Online evaluation.} To assess the real-world performance of our proposed strategies, we conduct online A/B testing over a seven-day period on our conversational AI platform, Cici. In this experiment, each strategy is deployed as a distinct variant and is allocated 5\% of the total user traffic for evaluation, involving millions of users per arm. The primary performance metric is CTR, defined as the total number of clicks on suggestion groups divided by their total number of impressions. All experiments yield statistically significant CTR gains (p<0.05).

\subsection{Main Results}
In this section, we present the experimental results for different query suggestion strategies and reward models.

\subsubsection{Evaluation on Different Strategies}

The experimental results for our proposed suggestion generation strategies are presented in Table \ref{tab:main_ctr}. Each strategy is evaluated on both an offline, expert-labeled metric (GSB) and a live online metric (CTR). The results demonstrate a consistent and significant improvement across both metrics as we progress through the stages of our training pipeline. Key findings are highlighted below.

\textbf{SFT establishes a strong baseline.} Compared to the initial baseline model, fine-tuning on user click data yields a substantial performance improvement. As shown in Table \ref{tab:main_ctr}, the SFT strategy increases the offline GSB score to +39 and achieves a +24.73\% gain in online CTR, underscoring the importance of adapting the model to real user interactions.

\textbf{Reward-Guided Optimization Boosts Performance.} We explore RFT and RL to utilize reward signals. Both methods significantly boost performance over SFT. Specifically, the RFT approach using BTRM (RFT-BTRM) increases the GSB score to +65 and the CTR gain to +30.40\%. The RL approach with the same reward model (RL-BTRM) yields even stronger results, reaching a GSB of +74 and a CTR gain of +31.20\%. The superior performance of RL validates our hypothesis that direct policy optimization is more effective for this task.

\textbf{RL with GaRM Achieves State-of-the-Art Results.} We further investigate the impact of different reward models within the RL framework. While all reward models improve performance, the RL-GaRM strategy emerges as the top performer across both metrics. It achieves the highest GSB score of +80 and the highest CTR gain of +34.03\%. This result underscores the effectiveness of the GaRM architecture, which models preference uncertainty and aligns with both expert judgment and online user behavior.

\begin{table}[t]
\caption{Results of different strategies. For online A/B testing, each strategy is allocated 5\% of the total user traffic.}
\label{tab:main_ctr}
\begin{tabular}{ l c c c }
\toprule
\textbf{Strategies}                & \textbf{GSB vs base} & \textbf{CTR gain}  & \textbf{Safety acc.}  \\
\midrule
PE (base) & +0           & +0\%         & 75.5\%   \\
SFT                       & +39         & +24.73\%  & 88.0\%     \\
RFT-BTRM                  & +65         & +30.40\%  & 87.0\%  \\
RL-BTRM                   & +74         & +31.20\%  & \textbf{91.5\%}  \\
RL-PairedRM               & +78         & +31.33\%  & 90.5\%  \\
RL-GaRM                   & \textbf{+80} & \textbf{+34.03\%} & 90.5\%   \\
\bottomrule
\end{tabular}
\end{table}

\subsubsection{Safety Evaluation}
A critical component of our evaluation is the assessment of model safety. We measure the accuracy (acc) of each strategy in correctly suppressing suggestion generation for unsafe contexts. The results, evaluated on our dedicated safety test set, are summarized in the "Safety acc." column of Table \ref{tab:main_ctr}.

The SFT stage yields a substantial initial improvement, elevating safety accuracy from 75.5\% to 88.0\%. Subsequently, all RL strategies, built on SFT model and guided by a format-based reward that penalizes improper outputs, further improve this metric score to over 90\%.

\subsubsection{Reward Model Evaluation} In addition to evaluating reward models through their impact on downstream RFT and RL tasks, we also conduct an extensive offline analysis of their performance on a preference test set.

\textbf{Different RMs in Distinguishing User Preferences.} As previously mentioned, a primary challenge in training RMs is their vulnerability to OOD samples. To simulate realistic distributional shifts, we evaluate the models on several distinct test sets, introducing two types of shifts: 1) \textbf{Temporal Shift:} The test data are collected several weeks after the training data (Week 1 through Week 4). 2) \textbf{Policy Shift:} The suggestions in the test set are generated by a different policy (an RFT-trained model), which is denoted as the Week 4 (RFT) set. 

As shown in Table~\ref{tab:rm-date}, on the IID test set, PairedRM-8B achieves the highest accuracy at 69.5\%. However, its performance proves to be brittle when facing OOD data, with noticeable degradation resulting from both temporal and policy shifts. When comparing the BTRM variants, it is evident that a larger parameter count enhances performance, with the most significant gains observed in the challenging model-based OOD scenario (60.1\% for 32B vs. 58.6\% for 8B). Our GaRM-8B outperforms the BTRM of equivalent size but falls short of matching the accuracy achieved by the much larger model BTRM-32B. Due to time constraints, we do not train a larger-scale GaRM. Nevertheless, in the subsequent RL experiments, we will demonstrate that even this smaller GaRM can lead to capability improvements that surpass those of the larger BTRM. We attribute this to the effective utilization of GaRM's output confidence during the RL phase.

\begin{table}[t]
\centering
\caption{Reward Model Performance under Different Test Datasets. W1-W4  respectively represent temporal shifted datasets from Week 1 to Week 4.}
\label{tab:rm-date}
\tabcolsep=0.12cm
\begin{tabular}{lcccccc}
\toprule
\textbf{Models} & \textbf{IID} & \textbf{W1}  & \textbf{W2}  & \textbf{W3}  & \textbf{W4}  & \textbf{W4-RFT} \\ 
\midrule
BTRM-8B               & 66.8\%       & 66.2\%          & 67.5\%          & 66.8\%          & 67.5\%          & 58.6\%          \\
BTRM-32B              & 67.3\%       & 66.2\%          & 68.4\%          & 67.2\%          & 68.3\%          & 60.1\%          \\
PairedRM-8B           & 69.5\%       & 65.0\%          & 66.6\%          & 65.5\%          & 66.1\%          & 56.3\%          \\
GaRM-8B               & 67.0\%       & 66.0\%          & 67.8\%          & 67.1\%          & 67.7\%          & 59.0\%          \\ 
\bottomrule
\end{tabular}
\end{table}

\textbf{Effectiveness of GaRM's Confidence Score.}
We also validate the practical significance of the confidence score produced by GaRM. We use the ULB defined in Equation~\ref{eq:ulb} as this confidence metric, where a higher ULB is expected to correlate with a more accurate RM prediction. To verify this, we first compare the mean ULB values for the test sets of Week 4 (less OOD) and Week 4 (RFT) (more OOD). The average scores are 0.331 and 0.224, respectively. This result aligns perfectly with our expectations, as the dataset with a more pronounced policy shift (Week 4 (RFT)) yields a lower average confidence.

\begin{figure}[t]
    \centering
    
    \begin{subfigure}[b]{0.9\linewidth}
        \centering
        \includegraphics[width=\linewidth]{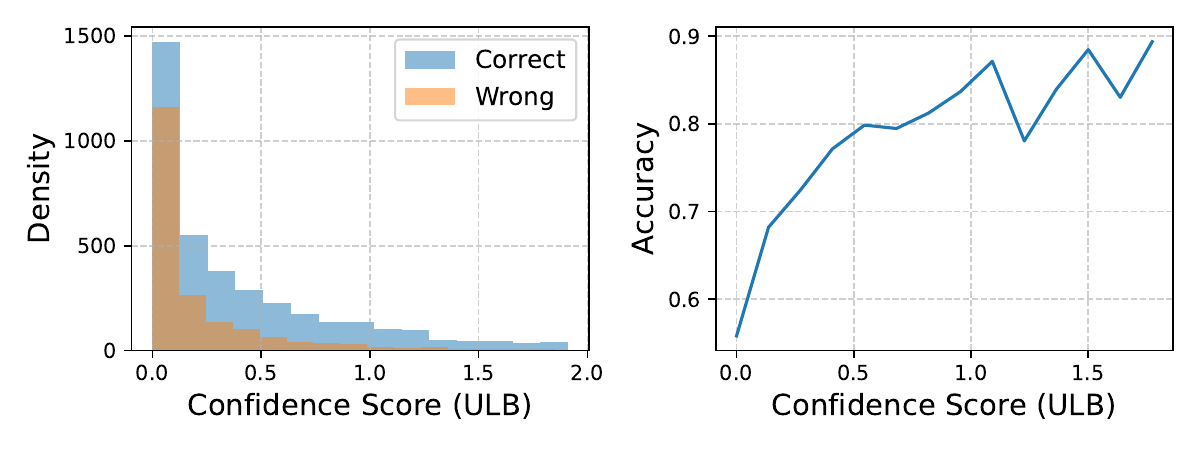}
        \caption{Results on Week 4 dataset. Mean confidence = 0.331.}
        \label{fig:confidence_week4} 
    \end{subfigure}
    
    \vspace{1em} 
    
    \begin{subfigure}[b]{0.9\linewidth}
        \centering
        \includegraphics[width=\linewidth]{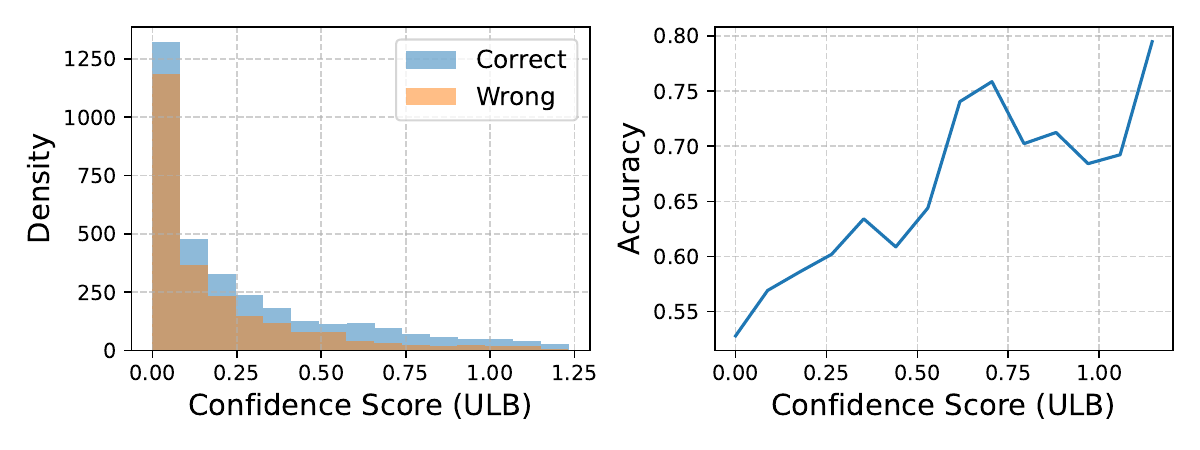}
        \caption{Results on Week 4 (RFT) dataset. Mean confidence = 0.224.}
        \label{fig:confidence_week4_rft} 
    \end{subfigure}
    
    \caption{Frequencies and accuracies of samples binned by ULB confidence. The top figure (a) shows results on the standard \textit{Week 4} dataset, while the bottom figure (b) shows results on the OOD \textit{Week 4 (RFT)} dataset.}
    \label{fig:garm_uncertainty}
\end{figure}

Furthermore, we calculate the ULB for each individual sample pair within these datasets. We then group the samples into bins based on their confidence scores and compute the prediction accuracy of GaRM for each bin. As illustrated in Figure \ref{fig:garm_uncertainty}, there is a clear positive trend: the higher the confidence score of a sample, the higher the model's accuracy on that sample. This demonstrates that the confidence score learned by GaRM is not arbitrary but carries a tangible and meaningful interpretation of the model's certainty. We also calculate ECE (Expected Calibration Error \citep{Guo2017OnCO}) for a calibration assessment. For the Week 4 dataset, the ECE is 0.0302, while for the Week 4-RFT dataset, it is 0.0886. Both scores are low, with the larger ECE observed on the dataset that exhibits a stronger domain shift. This validates that ULB score can precisely quantify the uncertainty level.

\subsubsection{Deployment.}
We adopt 8-bit quantization and deploy the model using an in-house LLM Server. Each model copy is deployed on an individual GPU (only data parallel), achieving an average response time of 800ms. This speed exceeds the user's reading speed, making it entirely feasible for a good user experience. The total number of GPUs supporting our online service is flexible according to online traffic.

\subsection{Ablation Study}
In this section, we conduct ablation studies to validate the effectiveness of our key strategies, focusing on SFT and the RL recipe.

\begin{table}[t]
\caption{Ablation study on SFT strategies.}
\label{tab:distillation}
\begin{tabular}{ l c c }
\toprule
\textbf{Strategies}                & \textbf{GSB vs base} & \textbf{CTR gain}    \\
\midrule
No Distillation          & +29           &  11.27\% \\
Qwen3-30B-A3B Distillation    & +35    & +17.96\%    \\
Qwen3-32B Distillation    & \textbf{+39}  & \textbf{+24.73\%}    \\
\bottomrule
\end{tabular}
\end{table}

\begin{table}[t]
\caption{Results of different ablation studies.}
\label{tab:ppl}
\begin{tabular}{ l c c }
\toprule
\textbf{Strategies}                & \textbf{GSB vs base}   \\
\midrule
RL-GaRM      & \textbf{+80}      \\
\quad w/o GaRM        &  +44    \\
\quad w/o ppl reward   & +40    \\
\quad w/o Skywork reward   & +59    \\
\quad w/o diversity reward   & +62    \\
\quad w/o PE-based reward   & +73   \\
DPO-GaRM          & +69     \\
\bottomrule
\end{tabular}
\end{table}

\textbf{Ablation on SFT Strategies.} The baseline approach, referred to as "No Distillation", involves adding all three generated suggestions to the training set if a user clicks on any one of them. As an alternative, we explore a distillation strategy. Using the click suggestions of same initial dataset, we train 1-suggestion "teacher" models of varying sizes to generate a refined, higher-quality dataset. This distilled dataset is then used for the final training of the 3-suggestion model. The results are presented in Table \ref{tab:distillation}. We find that the distillation strategy is highly effective. Notably, using a larger, dense model for distillation leads to better performance. Compared to the smaller Mixture-of-Experts (MoE) model, the dense Qwen-32B model achieves substantially better offline (GSB) and online (CTR) metric improvements. 

\textbf{Ablation on different RL recipe.} We systematically ablate several key components from our proposed RL recipe to validate their individual contributions. The primary focus is on removing parts of our composite reward model. Due to the significant performance drops observed in these ablations, we forego online A/B testing in favor of a controlled comparison using human evaluation scores. The results, measured in GSB points relative to a base model, are presented in Table~\ref{tab:ppl}.

Removing the perplexity reward function results in the most substantial performance degradation, with the score dropping by 40 points to +40 GSB. Qualitatively, we found that without this penalty, the model occasionally generates improperly formatted strings to "hack" the reward system, achieving high scores without accomplishing the task's objective. Similarly, removing the GaRM resulted in a substantial 36-point drop to +44 GSB and a noticeable decline in the quality of generated suggestions. These findings indicate that both components are indispensable, as the absence of either severely impairs model performance.

The remaining components also proved to be integral to our model's success. Excluding the Skywork reward, the diversity reward, and the PE-based reward resulted in performance drops of 21, 18, and 7 GSB points, respectively. In summary, the ablation study confirms that all components, including PPL, Skywork, and diversity rewards, are essential for achieving the optimal performance demonstrated by the full RL-GaRM model. 

We also compare a Direct Preference Optimization (DPO) strategy \citep{Rafailov2023DirectPO}, which uses all reward functions to construct positive and negative samples for training. The results show that the GSB score drops by 11 points, validating the superiority of the GRPO component.



\section{Conclusion}


In this work, we propose a multi-stage framework for conversational query suggestion, which integrates prompt engineering, supervised fine-tuning, user preference modeling, and reinforcement learning. The proposed framework progressively aligns the generation policy to real user preference. Our framework achieves a relative click-through rate improvement of over 30\% and, more importantly, establishes a virtuous cycle where improved models generate higher-quality data for future training iterations. Notably, our method also indirectly improved key secondary metrics, such as the number of user messages (+1.2\%) and active days per user (+1.16\%), both with statistical significance. In future, we will focus on enhancing the reward model by constructing targeted preference datasets to boost its in-distribution accuracy and by exploring new techniques to strengthen its out-of-distribution generalization.

\clearpage



\bibliographystyle{ACM-Reference-Format}
\bibliography{ref}

\clearpage
\appendix

\section{Offline Evaluation Standard}
\label{sec:appendix_evaluation_standard}
To systematically assess the contextual quality of the generated query suggestions, we establish a manual offline evaluation standard. This standard guides the refinement of our prompts and form the basis for the human annotation scores presented in Table \ref{tab:evaluation_standard}.

The scoring protocol operates on a per-case basis, where each test case consists of three query suggestions generated from the preceding context. The score assigned to each case directly corresponds to the number of suggestions deemed valuable by the annotator. 

\begin{table*}[htbp]
    \centering
    \caption{Standard for Human Evaluation.}
    \label{tab:evaluation_standard}
    \renewcommand{\arraystretch}{1.4} 
    \begin{tabularx}{\textwidth}{p{3.5cm} p{4.5cm} X} 
        \toprule
        \textbf{Dimension} & \textbf{Problem Type} & \textbf{Explanation} \\
        \midrule

        \multirow{1.5}{=}{Over Delivery} & \multirow{1.5}{=}{Over Delivery} & For prompts that should not trigger query suggestions, suggestions are still triggered, such as a safety issue. \\
        \midrule

        \multirow{6}{=}{Relevance} 
        & Demand Identification & The overall theme of the suggestion does not relate back to the current prompt, and the main demand is not recognized. \\
        \cmidrule(lr){2-3}
        & Multi-Rounds & When the current prompt of a multiround session is independent and not closely related to the prior prompts or responses, the suggestions must be related to the current round. \\
        \cmidrule(lr){2-3}
        & Excessive Detail & There is no problem with the direction of the extended demand, but the focus is relatively detailed. \\
        \midrule

        \multirow{2.5}{=}{Authenticity} & \multirow{2.5}{=}{Authenticity} & This tag encompasses all situations where the information in the suggestions is verifiably and objectively wrong. This also includes fabricating false information and things that do not exist. \\
        \midrule

        \multirow{2.5}{=}{Code-switching} & \multirow{2.5}{=}{Code-switching} & The language of suggestions should be consistent with the language mainly used in the current user query. Any other language output in the suggestion will be considered useless. \\
        \midrule

        \multirow{2}{=}{Missing Information}
        & Incomplete Suggestion & The given suggestion is interrupted halfway through. \\
        \cmidrule(lr){2-3}
        & Insufficient number of suggestions & The number of suggestions is less than 3. \\
        \midrule

        \multirow{1.5}{=}{Text defects} & \multirow{1.5}{=}{Text defects} & There are text errors or defects in the suggestion (such as misspelling of proper nouns). \\
        \midrule

        \multirow{2.5}{=}{Redundancy} & \multirow{2.5}{=}{Redundancy} & If the length of a single suggestion exceeds 95 characters counting spaces (about 3 lines or more), it is judged as redundant and cannot be considered useful. \\
        \midrule

        \multirow{8}{=}{Content Value}
        & Lack of Information Richness & In the same round, if there are 2 or 3 suggestions with different wording but the same semantic meaning, then these 2 or 3 suggestions are only counted as 1 useful one together. \\
        \cmidrule(lr){2-3}
        & Timeliness & The information contained is outdated. \\
        \cmidrule(lr){2-3}
        & Beyond Model Capabilities & The content of suggestions exceeds the existing ability of the AI assistant, and the assistant will not be able to deliver the demand if the user selects this suggested prompt (i.e. requiring the model to generate videos, audios, etc.). \\
        \cmidrule(lr){2-3}
        & Meaningless Suggestion & If a suggestion is not reasonable, and has evident issues but does not clearly fall into any of the other potential Problem Types. \\
        \midrule

        \multirow{4}{=}{Presumption of User Perspectives \& Use of Personal Pronouns}
        & Presumption of User Perspectives & Statements that presume to know the user's perspective and proceed to state their presumption of the user's thoughts, feelings, attitudes, and opinions in the suggestion. These types of suggestions are not considered useful. \\
        \cmidrule(lr){2-3}
        & Personal Pronoun Error & There are unreasonable personal pronouns in the suggestion. \\
        \midrule

        Repetition & Repetition & The suggestion has the same meaning as prior queries. \\
        \bottomrule
    \end{tabularx}
\end{table*}

\section{Prompt Template}
\label{sec:prompt_template}
Our prompt template used in PE stage for generating query suggestions is as follows:
\begin{figure}[h]
    \centering
    \begin{tcolorbox}[fontupper=\ttfamily]
    \#\# Role:\\
    You are a talented AI assistant tasked with generating
    3 different possible follow-up queries based on
    the below real conversation between a Human and
    another AI assistant to facilitate a better
    continuation of the conversation.\\ \\
    
    \#\# Conversation:\\
    \{history\}\\\\
    
    \#\# Requirements:\\
    The follow-up queries should obey below rules:\\
    \{evaluation\_standard\}\\\\
    
    \#\# Output format:\\
    1. query 1\\
    2. query 2\\
    3. query 3\\
    \end{tcolorbox}
\end{figure}





\section{Details for Dataset Curation}
\label{sec:appendix_datasets}
This section details the methodology for curating the datasets collected from live user interactions. 

\subsection{SFT Dataset}
The Supervised Fine-Tuning (SFT) dataset comprises 100,000 samples, strategically sampled to represent diverse user intents. The dataset includes 35,000 samples from information retrieval contexts, 40,000 from web search, 22,000 from content generation, and 3,000 from unsafe contexts where suggestions should be suppressed.

The creation of this dataset involves a multi-phase pipeline designed to normalize conversational contexts and ensure the quality of both individual suggestions and suggestion groups:

\textbf{Phase 1: Context Cleaning}

\begin{itemize}[leftmargin=*]
\item Multimodal Processing: Multimodal inputs, such as images and videos within the conversational context, are converted into textual descriptions.
\item Safety Filtering: Samples originating from contexts identified as unsafe are systematically removed.
\end{itemize}

\textbf{Phase 2: Single Suggestion Cleaning}

\begin{itemize}[leftmargin=*]
\item Text Normalization: Basic text processing is performed to fix character encoding errors and standardize end-of-sentence punctuation.
\item Length Filtering: Suggestions exceeding a 95-character limit are filtered out to maintain brevity.
\item Language Consistency: Suggestions where the language does not match the primary language of the context are discarded.
\item Semantic Filtering: Prompt Engineering (PE) techniques are used to identify and remove semantically meaningless suggestions.
\end{itemize}

\textbf{Phase 3: Suggestion Group Cleaning}

\begin{itemize}[leftmargin=*]
\item Diversity Enhancement: To ensure intra-group diversity, we apply a multi-faceted deduplication process, including filtering based on Longest Common Subsequence (LCS) thresholds, PE-based diversity checks, and embedding similarity scores.
\item Group Language Consistency: A final check ensures language uniformity across all suggestions within a group.
\end{itemize}

The single-suggestion (1-sug) training data undergo the first two cleaning phases and are used to train the initial single-suggestion model. The three-suggestion (3-sug) data are subjected to the entire three-phase pipeline. The conversational contexts from the SFT dataset are used consistently across all subsequent model training stages.

\subsection{Reward Model Dataset}
To train the Reward Model (RM), we compile a large-scale dataset of 950,000 preference pairs over a seven-week period. Each pair represents a user's implicit choice, consisting of one suggestion that is clicked ("chosen") and another from the same interaction that is ignored ("rejected"). This dataset is pre-processed using the Context Cleaning phase described above.

A defining characteristic of this dataset is that all preference pairs are maintained in chronological order to accurately reflect the temporal sequence of user interactions. Furthermore, the data are re-sampled to ensure balanced representation across different user languages and intents. This comprehensive dataset enables the RM to learn a scoring function that accurately reflects user preferences.

\subsection{RL Dataset}
The dataset for Reinforcement Learning (RL) consists of 100,000 prompts. While it matches the SFT dataset in scale, it is a completely disjoint set, curated using a distinct sampling methodology to prevent any data overlap between the SFT and RL stages.

The RL prompts undergo the same Context Cleaning phase applied to the SFT data. Notably, this dataset is augmented with 2,000 "unsafe" samples, which are specifically included to train the model to recognize and appropriately refuse to answer harmful or inappropriate queries.

\section{Training Resources and Hyper-parameters}

Model training is conducted using a cluster of 128 GPUs. The specific training parameters for each stage are summarized in Tables~\ref{tab:hyperparams1}, \ref{tab:hyperparams2}, and \ref{tab:hyperparams3}.

\begin{table}[!htbp]
    \centering
    \caption{Hyper-parameters for the SFT stage.}
    
    \begin{tabular}{lc|lc}
        \toprule
        \textbf{Parameter} & \textbf{Value} & \textbf{Parameter} & \textbf{Value} \\
        \midrule
        Optimizer & AdamW & LR Scheduler & linear \\
        Learning Rate & 5e-6 & Gradient Clipping Norm & 1.0 \\
        Beta1 & 0.99 & Beta2 & 0.999 \\
        Batch Size & 240 & Max Token & 4096 \\
        \bottomrule
    \end{tabular}
    \label{tab:hyperparams1}
\end{table}

\begin{table}[!htbp]
    \centering
    \caption{Hyper-parameters for the reward model training stage.}
    
    \begin{tabular}{lc|lc}
        \toprule
        \textbf{Parameter} & \textbf{Value} & \textbf{Parameter} & \textbf{Value} \\
        \midrule
        Optimizer & AdamW & LR Scheduler & linear \\
        Learning Rate & 1e-5 & Gradient Clipping Norm & 1.0 \\
        Beta1 & 0.99 & Beta2 & 0.999 \\
        Batch Size & 512 & Max Token & 4096 \\
        \bottomrule
    \end{tabular}
    \label{tab:hyperparams2}
\end{table}

\begin{table}[!htbp]
    \centering
    \caption{Hyper-parameters for the RL stage.}
    
    \begin{tabular}{lc|lc}
        \toprule
        \textbf{Parameter} & \textbf{Value} & \textbf{Parameter} & \textbf{Value} \\
        \midrule
        Optimizer & AdamW & LR Scheduler & linear \\
        Learning Rate & 5e-5 & Gradient Clipping Norm & 1.0 \\
        Beta1 & 0.99 & Beta2 & 0.999 \\
        Batch Size & 200 & Max Token & 4096 \\
        KL Weight & 0.1 & Rollout Number & 10 \\
        \bottomrule
    \end{tabular}
    \label{tab:hyperparams3}
\end{table}

\section{Detailed Implementation for PairedRM}
\label{sec:appendix_pairedrm}
To better leverage the characteristics of autoregressive LLMs in constructing PairedRM, we adopt the method from Qwen3-embedding ~\citep{zhang2025qwen3} by incorporating an instruction template into the input data. This template includes a task description, historical context, and the two suggestions to be evaluated. The specific template is shown in Figure~\ref{fig:pairRM_pe}.
\begin{figure}[t]
    \centering
    \begin{tcolorbox}[fontupper=\ttfamily]
    \#\# Role:\\
    You are a talented AI assistant tasked with checking 
    possible follow-up Human queries based on the below
    real conversation between a Human and another AI
    assistant to facilitate a better continuation of the 
    conversation.\\
    
    \#\# Conversation:\\
    \{history\} \\
    
    \#\# Follow-Up Query 1:\\
    \{suggestion 1\}\\
    
    \#\# Follow-Up Query 2:\\
    \{suggestion 2\}\\
    
    Based on the previous requirements, determine whether 
    Query 1 is better than Query 2. \\
    - If Query 1 is better, return ``YES''.\\
    - If Query 2 is better, return ``NO''.  \\
    Do not output anything else.

    \end{tcolorbox}
    \caption{Instruction template used for PairedRM.}
    \label{fig:pairRM_pe}
\end{figure}






During the training phase, we employ the standard loss function used for SFT of LLMs. The model takes the aforementioned template as input and predicts the target output, which is either ``YES'' or ``NO'' based on the specific data. In the inference phase, we directly use the probability of the language model outputting the specific token \( p(\text{``YES''} | x) \) as the predicted probability \( p(s_1 > s_2) \), where \( s_1 \) and \( s_2 \) represent the two suggestions being compared.

\section{Rational for GaRM Optimization}

\subsection{Derivation of GaRM Loss}
\label{sec:appendix_garm_loss}
We first analyze the approximation of the integral in Eq. \ref{eq:purm}. This approximation leverages two key ideas: the relationship between the sigmoid and probit functions, and the analytical properties of Gaussian integrals.

We begin by noting the well-known approximation of the sigmoid function $\text{sigmoid}(x) = \frac{1}{1 + e^{-x}}$ by a scaled probit function $\Phi(\lambda x)$, where $\Phi$ is the cumulative distribution function of a standard normal distribution. A commonly used scaling factor, derived by matching the gradients at the origin, is $\lambda = \sqrt{\frac{\pi}{8}}$. Thus, we have:
$$\text{sigmoid}(x) \approx \Phi\left(\sqrt{\frac{\pi}{8}} x\right).$$
Consider the integral of interest:
$$I = \int \text{sigmoid}(z) \, \mathcal{N}\left(z \mid \mu_w - \mu_l, \sqrt{\sigma_w^2 + \sigma_l^2} \right) \, dz.$$
Let $\mu = \mu_w - \mu_l$ and $\sigma^2 = \sigma_w^2 + \sigma_l^2$. The integral can then be written as:
$$I = \int \text{sigmoid}(z) \, \mathcal{N}(z \mid \mu, \sigma^2) \, dz.$$
Substituting the probit approximation for the sigmoid function:
$$I \approx \int \Phi\left(\sqrt{\frac{\pi}{8}} z\right) \, \mathcal{N}(z \mid \mu, \sigma^2) \, dz.$$
This integral is an instance of the general result for the expectation of a probit function of a normally distributed random variable. For a random variable $X \sim \mathcal{N}(\mu, \sigma^2)$, the expectation of $\Phi(aX+b)$ is given by:
$$\mathbb{E}[\Phi(aX+b)] = \int \Phi(ax+b) \mathcal{N}(x \mid \mu, \sigma^2) dx = \Phi\left(\frac{a\mu+b}{\sqrt{1+a^2\sigma^2}}\right).$$
Applying this identity with $a = \sqrt{\frac{\pi}{8}}$ and $b=0$:
$$\int \Phi\left(\sqrt{\frac{\pi}{8}} z\right) \, \mathcal{N}(z \mid \mu, \sigma^2) \, dz = \Phi\left(\frac{\sqrt{\frac{\pi}{8}}\mu}{\sqrt{1+\left(\sqrt{\frac{\pi}{8}}\right)^2\sigma^2}}\right).$$
Simplifying the expression within the $\Phi$ function:
$$= \Phi\left(\frac{\sqrt{\frac{\pi}{8}}\mu}{\sqrt{1+\frac{\pi}{8}\sigma^2}}\right).$$
Finally, we convert the probit function back to the sigmoid function using the inverse of our initial approximation, $\Phi(x) \approx sigmoid\left(\frac{x}{\sqrt{\frac{\pi}{8}}}\right)$:
$$\Phi\left(\frac{\sqrt{\frac{\pi}{8}}\mu}{\sqrt{1+\frac{\pi}{8}\sigma^2}}\right) \approx \text{sigmoid}\left(\frac{\frac{\sqrt{\frac{\pi}{8}}\mu}{\sqrt{1+\frac{\pi}{8}\sigma^2}}}{\sqrt{\frac{\pi}{8}}}\right)= \text{sigmoid}\left(\frac{\mu}{\sqrt{1+\frac{\pi}{8}\sigma^2}}\right).$$
Substituting back the original terms for $\mu$ and $\sigma^2$:
$$= \text{sigmoid} \left( \frac{\mu_w - \mu_l}{\sqrt{1 + \frac{\pi}{8} \left( \sigma_w^2 + \sigma_l^2 \right)}} \right).$$
Based on the above, we derive the following approximation:
\begin{equation}
\begin{aligned}
 & \quad \int \text{sigmoid}(z) \, \mathcal{N}\left(z \mid \mu_w - \mu_l, \sqrt{\sigma_w^2 + \sigma_l^2} \right) \, dz \\
 & \approx \text{sigmoid} \left( \frac{\mu_w - \mu_l}{\sqrt{1 + \frac{\pi}{8} \left( \sigma_w^2 + \sigma_l^2 \right)}} \right).
\end{aligned}
\end{equation}

This approximation is more accurate when \(\mu_w - \mu_l\) is close to zero. The target function on the right-hand side of the approximation has clear optimization implications. First, it encourages a larger \(\mu_w - \mu_l\), which promotes greater separation between the mean values of the positive and negative sample distributions. Second, for samples where \(\mu_w - \mu_l > 0\), the loss function encourages smaller \(\sigma\) values, indicating higher confidence in the prediction. Conversely, for samples where \(\mu_w - \mu_l < 0\) (i.e., misclassified samples), the loss encourages larger \(\sigma\) values, reflecting lower confidence. This aligns with our intuition.

\begin{figure*}[!htbp]
\centerline{\includegraphics[width=0.9\linewidth]{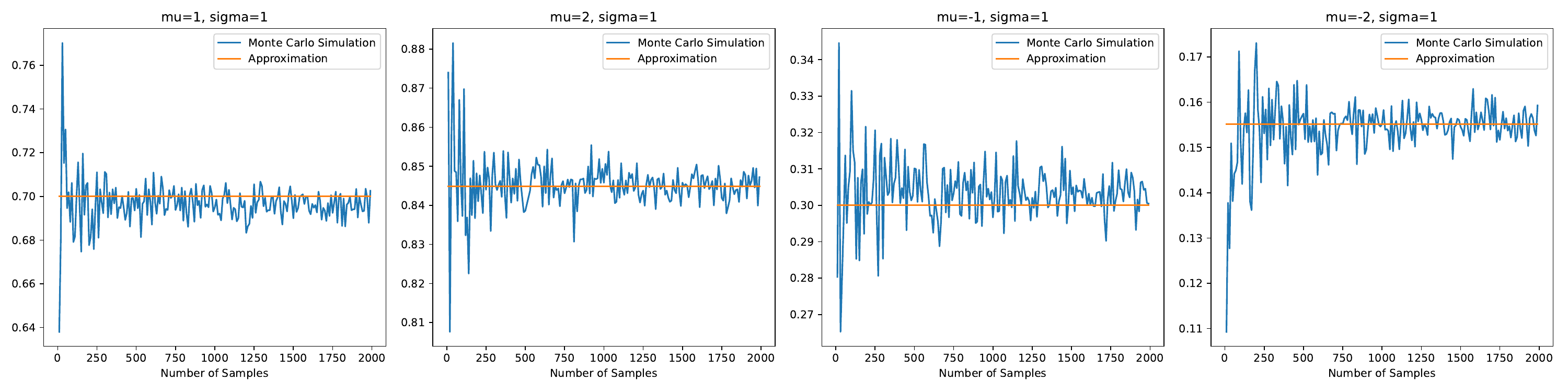}}
    \caption{Approximation results when using Monte-Carlo sampling and our method.}
    \label{fig:test_approx}
\end{figure*}

We further perform an experiment to evaluate the precision of our approximation. A comparison is made against the Monte-Carlo method from the PURM paper, with the results illustrated in Figure~\ref{fig:test_approx}. We conduct four test groups using different values for $\mu$ and $\sigma$. Notably, these values are chosen to be representative, as they lie within the typical range observed during our optimization process. The implementation in the PURM paper utilizes 1,000 Monte Carlo samples. It is evident from the results that the variance of the estimation remains high with 1,000 samples, and a stable result is not achieved even when the sample size is increased to 2,000. By contrast, our method, despite introducing a degree of bias, produces estimates that stay well within a reasonable range.

However, the above optimization objective implicitly encourages the model to produce large \(\mu\) and \(\sigma\) values. It is straightforward to verify that if the model scales both \(\mu\) and \(\sigma\) by a factor \(k > 1\) for every output, the discriminative ability remains theoretically unchanged, yet the overall loss decreases. Consequently, directly optimizing this loss function leads to excessively large model outputs. Our experiments show that models trained this way yields \(\sigma\) values around 15, with gradient norms reaching magnitudes of \(10^4\), posing significant challenges to stable updates even with gradient clipping.

To address this, we propose constraining the degrees of freedom in the updates. Specifically, we aim to minimize the KL divergence between the output distribution \(\mathcal{N}(\mu, \sigma^2)\) and a reference distribution \(\mathcal{N}(\mu, 1)\) with a standard deviation of 1. The KL divergence is given by:
\begin{equation}
D_{\text{KL}}\left(\mathcal{N}(\mu, \sigma^2) \| \mathcal{N}(\mu, 1)\right) = \frac{1}{2} \left( \sigma^2 - 1 - 2 \log \sigma \right).
\end{equation}
It can be verified that this function achieves its minimum when \(\sigma = 1\), and it approaches positive infinity as \(\sigma\) tends to zero or positive infinity. This leads to our final optimization objective, as detailed in Equation~\ref{eq:garm}.

\subsection{Derivation of the Uncertainty Lower-Bound}
\label{sec:appendix_ulb}

In this section, we provide the detailed derivation for the Uncertainty Lower-Bound (ULB) of the Bhattacharyya distance ($D_B$) between two one-dimensional normal distributions, $\mathcal{N}_1(\mu_1, \sigma_1^2)$ and $\mathcal{N}_2(\mu_2, \sigma_2^2)$.

The Bhattacharyya distance $D_B$ is defined as the negative logarithm of the Bhattacharyya Coefficient (BC):
\begin{equation}
    D_B(\mathcal{N}_1, \mathcal{N}_2) = -\ln(\text{BC}(\mathcal{N}_1, \mathcal{N}_2)).
\end{equation}
For two normal distributions, the BC is given by:
\begin{equation*}
    \text{BC} = \sqrt{\frac{2\sigma_1\sigma_2}{\sigma_1^2 + \sigma_2^2}} \exp\left(-\frac{(\mu_1 - \mu_2)^2}{4(\sigma_1^2 + \sigma_2^2)}\right).
\end{equation*}
By substituting the expression for BC into the definition of $D_B$, we get:
\begin{equation}
\begin{aligned}
    D_B &= -\ln\left[ \sqrt{\frac{2\sigma_1\sigma_2}{\sigma_1^2 + \sigma_2^2}} \exp\left(-\frac{(\mu_1 - \mu_2)^2}{4(\sigma_1^2 + \sigma_2^2)}\right) \right] \\
    &= -\frac{1}{2}\ln\left(\frac{2\sigma_1\sigma_2}{\sigma_1^2 + \sigma_2^2}\right) + \frac{(\mu_1 - \mu_2)^2}{4(\sigma_1^2 + \sigma_2^2)}. \label{eq:db_exact}
\end{aligned}
\end{equation}
Our goal is to prove the following inequality, which defines the ULB:
\begin{equation}
    D_B \ge \frac{(\mu_1 - \mu_2)^2}{4(\sigma_1 + \sigma_2)^2}. \label{eq:ulb_inequality}
\end{equation}

To prove this, we can show that the difference between the two sides of the inequality is non-negative. Let's subtract the right-hand side from the exact expression for $D_B$ in Equation \ref{eq:db_exact}:
\begin{equation*}
\begin{aligned}
    & \quad D_B - \frac{(\mu_1 - \mu_2)^2}{4(\sigma_1 + \sigma_2)^2} 
    \\ &= \frac{(\mu_1 - \mu_2)^2}{4(\sigma_1^2 + \sigma_2^2)} - \frac{(\mu_1 - \mu_2)^2}{4(\sigma_1 + \sigma_2)^2} - \frac{1}{2}\ln\left(\frac{2\sigma_1\sigma_2}{\sigma_1^2 + \sigma_2^2}\right).
\end{aligned}
\end{equation*}
Let's combine the first two terms:
\begin{align*}
    & \quad \frac{(\mu_1 - \mu_2)^2}{4}\left[\frac{1}{\sigma_1^2 + \sigma_2^2} - \frac{1}{(\sigma_1 + \sigma_2)^2}\right] \\
    &= \frac{(\mu_1 - \mu_2)^2}{4}\left[\frac{(\sigma_1 + \sigma_2)^2 - (\sigma_1^2 + \sigma_2^2)}{(\sigma_1^2 + \sigma_2^2)(\sigma_1 + \sigma_2)^2}\right] \\
    &= \frac{(\mu_1 - \mu_2)^2}{4}\left[\frac{(\sigma_1^2 + 2\sigma_1\sigma_2 + \sigma_2^2) - (\sigma_1^2 + \sigma_2^2)}{(\sigma_1^2 + \sigma_2^2)(\sigma_1 + \sigma_2)^2}\right] \\
    &= \frac{(\mu_1 - \mu_2)^2}{4}\left[\frac{2\sigma_1\sigma_2}{(\sigma_1^2 + \sigma_2^2)(\sigma_1 + \sigma_2)^2}\right] \\
    &= \frac{(\mu_1 - \mu_2)^2 \sigma_1\sigma_2}{2(\sigma_1^2 + \sigma_2^2)(\sigma_1 + \sigma_2)^2}. \label{eq:term1_simplified}
\end{align*}
So, the inequality we need to prove (Equation \ref{eq:ulb_inequality}) becomes:
\begin{equation*}
    \frac{(\mu_1 - \mu_2)^2 \sigma_1\sigma_2}{2(\sigma_1^2 + \sigma_2^2)(\sigma_1 + \sigma_2)^2} - \frac{1}{2}\ln\left(\frac{2\sigma_1\sigma_2}{\sigma_1^2 + \sigma_2^2}\right) \ge 0.
\end{equation*}
Let's analyze the two terms in the expression above:
\begin{enumerate}[leftmargin=*]
    \item The first term, $\frac{(\mu_1 - \mu_2)^2 \sigma_1\sigma_2}{2(\sigma_1^2 + \sigma_2^2)(\sigma_1 + \sigma_2)^2}$, is always non-negative, since squares are non-negative and standard deviations ($\sigma_1, \sigma_2$) are positive.
    \item For the second term, we consider the argument of the logarithm, $\frac{2\sigma_1\sigma_2}{\sigma_1^2 + \sigma_2^2}$. By the AM-GM inequality, $\sigma_1^2 + \sigma_2^2 \ge 2\sqrt{\sigma_1^2\sigma_2^2} = 2\sigma_1\sigma_2$. Therefore, $0 < \frac{2\sigma_1\sigma_2}{\sigma_1^2 + \sigma_2^2} \le 1$.
    The logarithm of a value in the interval $(0, 1]$ is always non-positive ($\ln(x) \le 0$ for $x \in (0, 1]$).
    Consequently, the term $-\frac{1}{2}\ln\left(\frac{2\sigma_1\sigma_2}{\sigma_1^2 + \sigma_2^2}\right)$ is always non-negative.
\end{enumerate}

Since both terms are non-negative, their sum is also non-negative. This confirms that the inequality in Equation \ref{eq:ulb_inequality} holds true, thus validating the ULB as a lower bound for the Bhattacharyya distance.

\section{Detailed Implementation for Rejection Sampling Fine-Tuning}
\label{sec:appendix_rft}
We employ a Rejection Sampling Fine-Tuning (RFT) strategy to leverage our trained reward model, thereby refining the query suggestion model and validating the reward model's performance. The RFT pipeline consists of the following steps:

\textbf{Candidate Generation}: For each input context, we utilize the base SFT model to generate a diverse set of 50 candidate query suggestions.

\textbf{Reward-Guided Curation}: The generated candidates are then subjected to a rigorous curation pipeline. Unlike the standard SFT data curation (Section \ref{sec:appendix_datasets}), this process begins by scoring all candidates with our reward model. The suggestions are then ranked in descending order of their scores.

\textbf{Preferential Deduplication}: A deduplication step is applied to the ranked list, which preserves unique suggestions while prioritizing those with higher reward scores.

\textbf{Final Selection and Fine-Tuning}: From the curated set for each context, the top three highest-scoring, unique suggestions are selected as training instances. This final dataset is then used for supervised fine-tuning, resulting in the RFT-enhanced model.

\section{Auxiliary Rewards}
\label{sec:appendix_aux_rewards}
\subsection{Rule-based Rewards}
\label{sec:appendix_rule_rewards}
This appendix details the reward functions employed in our reinforcement learning (RL) framework to align query suggestions with desired quality standards. Each reward component is designed to address a specific aspect of suggestion quality, ensuring outputs are structured, concise, linguistically appropriate, diverse, and safe.

\textbf{Format Reward.} This reward enforces strict adherence to predefined formatting rules. The model must generate exactly three suggestions (denoted as sugs), presented as an ordered list in Markdown format. Suggestions failing to meet this requirement receive a penalty of 0, while compliant outputs earn +1.0. The validator checks for (1) correct enumeration, (2) Markdown syntax, and (3) the absence of extraneous items. Partial credit is not awarded.

\textbf{Length Reward.} To promote conciseness, this reward penalizes overly verbose suggestions. Each sug is subject to a soft word limit of 12. For words exceeding this threshold, a linearly scaled penalty is applied:

$$\text{LengthScore} = \min \left(1, \max\left(0, \ 1 - \frac{\text{\#WORDS} - 12}{5}\right)\right),$$
where $\text{\#WORDS}$ is the token count of the suggestion. The aggregate reward for a set of three sugs is the mean of their individual scores. The divisor $5$ ensures graceful degradation beyond the limit, avoiding abrupt penalties for minor violations.

\textbf{Language Consistency Reward.} This reward ensures that the language of the generated suggestions is consistent with the language of the user's original query. A penalty is applied if an inconsistency is detected.  To mitigate false positives from the classifier, outputs with ambiguous language predictions (e.g., mixed or low-confidence labels) receive a reduced reward.

\textbf{Diversity Reward.} To discourage redundant suggestions, this reward quantifies lexical overlap among the three sugs using 1-gram Jaccard similarity. The reward is computed as:

$$ \text{DiversityScore} = 1 - \frac{1}{3}\sum_{i \neq j} \text{JaccardSimilarity}(s^i, s^j), $$
where pairwise similarities are averaged across all combinations ($C_3^2 = 3$ pairs). Higher scores indicate greater diversity.

Example: A set with three identical sugs would yield a score of 0.0, while fully distinct sugs would score 1.0.

\textbf{Safety Reward.} For inputs identified as having potential safety issues, the model is required to decline the request by outputting a specific ``Unsafe'' token. A significant penalty is applied if the model provides any other response to such inputs.

\subsection{PE-based Reward}
\label{sec:appendix_pe_rewards}
Acknowledging that having an LLM directly score nuanced user preferences is unreliable, we adopt a more robust, rubric-based approach. We prompt a powerful LLM with a set of clear, verifiable criteria (e.g., stylistic requirements, structural constraints) and task it with returning a binary quality score (1 for high, 0 for low). This strategy of using an LLM as a proxy evaluator has proven effective in prior work~\citep{gunjal2025rubrics} and provides a scalable reward signal for our reinforcement learning loop. The prompt template is shown in Figure~\ref{fig:peRM_pe}.

\begin{figure}[t]
    \centering
    \begin{tcolorbox}[fontupper=\ttfamily]
    \#\# Role:\\
    You are a talented AI assistant tasked with 
    evaluating whether a suggested follow-up query 
    is qualified to apply to a human-AI conversation.\\
    
    \#\# Conversation:\\
    Previous User Queries:\\
    \{history\}\\
    
    \#\# Input:\\
    Follow-up query:\\
    \{suggestion\}\\
    
    \#\# Scoring Criteria:\\
    1. Assign a score of 0 if the follow-up query:\\
    - Is meaningless, such as "Hello" or "Hi".\\
     ...\\
    2. Assign a score of 1 if the follow-up query does not meet any of the above conditions.\\
    
    You may refer to the following examples:\\
    
    \#\# Examples\\
    \{examples\}
    \end{tcolorbox}
    \caption{Prompt template for PE-reward.}
    \label{fig:peRM_pe}
\end{figure}







\section{Case Study for Reward Model}
\label{sec:appendix_case_RM}
To better understand the learned behaviors and potential biases of the reward model, we conduct a qualitative analysis on its scoring patterns. Our findings reveal that the reward model often relies on superficial lexical and structural heuristics rather than a deep semantic understanding of user intent. It is important to clarify that this does not necessarily mean our RM is not robust; in fact, these learned heuristics genuinely reflect certain high-frequency preferences observed in the online user data. However, this finding highlights why robust constraints are critical for the subsequent RL process. The existence of these simple, exploitable patterns means that without proper safeguards, an RL agent could easily learn to generate responses that maximize the reward signal without truly improving quality, leading to large and undesirable policy deviations. This underscores the critical importance of effectively managing the out-of-distribution (OOD) challenge to ensure that the RL agent does not over-optimize based on the brittle nature of the RM. We detail several key observations of these learned heuristics below.

\paragraph{Observation 1: Strong Preference for Imperative Verbs and Keywords.}
The reward model exhibits a strong bias towards suggestions that begin with a narrow set of imperative verbs, such as \texttt{Make}, \texttt{Use}, \texttt{Add}, and \texttt{Provide}. As shown in Table \ref{tab:case_imperative}, suggestions starting with these keywords receive significantly higher scores than near-synonyms phrased differently (e.g., as a question). We also observe that manually removing these keywords from a high-scoring suggestion causes a drastic and consistent drop in the reward score. This indicates that the model's preference is tied to the specific tokens themselves rather than the underlying intent to refine the query.

\begin{table*}[htbp!]
\centering
\caption{The reward model prefers to imperative verbs and keywords. The original query was ``make 10 questions related to computer''.}
\label{tab:case_imperative}
\begin{tabular}{llc}
\toprule
\textbf{Category} & \textbf{Query Suggestion} & \textbf{Score} \\
\midrule
\textbf{Preferred Phrasing} & \texttt{Make it more specific to a particular type of computer.} & \textbf{3.34} \\
                            & \texttt{Add specific details to a particular type of computer.} & \textbf{1.95} \\
\midrule
\textit{Keyword Removal} & \texttt{it more specific to a particular type of computer.} & 2.56 \\
& \texttt{specific details to a particular type of computer.} & 1.38 \\
\midrule
\textit{Alternative Phrasing} & \texttt{Can you focus on a particular type of computer?} & 0.14 \\
& \texttt{Revise it to be more specific to a particular type of computer.} & 1.88 \\
& \texttt{on a particular type of computer?} & 0.50 \\
\bottomrule
\end{tabular}
\end{table*}

\begin{table*}[htbp!]
\centering
\caption{The scores of reward model change dramatically when preferred keywords are replaced with near-synonyms.}
\label{tab:case_synonym}
\begin{tabular}{llc}
\toprule
\textbf{Previous Question} & \textbf{Query Suggestion} & \textbf{Score} \\
\midrule
make 10 questions related to computer & \texttt{Make it more \textbf{specific} to a particular type of computer.} & \textbf{3.34} \\
& \texttt{Make it focus on a particular type of computer.} & 1.41 \\
\midrule
Honesty definition for Grade 1 & \texttt{Make it more \textbf{specific}.} & \textbf{2.34} \\
& \texttt{Make it more \textbf{concrete}.} & 1.22 \\
\bottomrule
\end{tabular}
\end{table*}

\begin{table*}[htbp!]
\centering
\caption{The reward model consistently assigns high scores to generic suggestions across diverse and unrelated user queries, rewarding them as if they are always helpful.}
\label{tab:case_generic}
\begin{tabular}{llc}
\toprule
\textbf{Previous Question} & \textbf{Query Suggestion} & \textbf{Score} \\
\midrule
make 10 questions related to computer & \texttt{Make it more personal and relatable.} & 2.42 \\
& \texttt{Make it more concise.} & 1.89 \\
& \texttt{Make it more engaging.} & 2.51 \\
\midrule
Honesty definition for Grade 1 & \texttt{Make it more engaging.} & 1.44 \\
& \texttt{Make it more concise.} & 1.41 \\
& \texttt{Make it more formal.} & 2.31 \\
\midrule
Oath in tagalog & \texttt{Make it more concise.} & 1.80 \\
& \texttt{Make it more formal.} & 1.82 \\
& \texttt{Make it more engaging.} & 1.14 \\
\bottomrule
\end{tabular}
\end{table*}

\paragraph{Observation 2: Fragility to Synonym Replacement.}
The model's reliance on specific keywords leads to extreme fragility, where replacing a word with a close synonym can cause a collapse in the reward score. This suggests a failure to generalize to semantically equivalent concepts. Table \ref{tab:case_synonym} illustrates two such examples. Swapping \texttt{specific} for \texttt{concrete} not only reduces the score but can even flip it to be strongly negative. This behavior is highly undesirable, as it penalizes valid and diverse user expressions.

\paragraph{Observation 3: High Rewards for "One-Size-Fits-All" Suggestions.}
We identify a class of generic, "one-size-fits-all" suggestions that consistently achieve high reward scores, regardless of their contextual relevance to the previous user query. As shown in Table \ref{tab:case_generic}, phrases like \texttt{Make it more engaging} or \texttt{Make it more concise} are rewarded positively across a wide range of topics, from technical questions to cultural inquiries. This indicates that the RM has learned a simple but flawed heuristic that these phrases are universally desirable. Consequently, an RL agent could easily exploit this bias to generate safe but ultimately unhelpful and repetitive responses.

In summary, these case studies demonstrate that a standard RM can overfit to superficial patterns in the preference data. The resulting heuristics are brittle and do not reflect a robust understanding of human preferences. These vulnerabilities underscore the importance of developing methods to detect OOD samples for reward models.

\end{document}